# Integrating Attendance Tracking and Emotion Detection for Enhanced Student Engagement in Smart Classrooms

Keith Ainebyona[1], Ann Move Oguti[2], Joseph Walusimbi[3], Ritah Kobusingye[4]

[1]keithainebyona14@gmail.com, [2]amoguti@sun.ac.ug, [3]mrjosephwalusimbi@gmail.com, [4]2401600224@sun.ac.ug

*Department of Electronics and Computer Engineering, School of Engineering and Technology, Soroti University*

The increasing adoption of smart classroom technologies in higher education has mainly focused on automating attendance, with limited attention given to students' emotional and cognitive engagement during lectures. This limits instructors' ability to identify disengagement and adapt teaching strategies in real time. This paper presents SCASED (Smart Classroom Attendance System with Emotion Detection), an IoT-based system that integrates automated attendance tracking with facial emotion recognition to support classroom engagement monitoring. The system uses a Raspberry Pi camera and OpenCV for face detection, and a fine-tuned MobileNetV2 model to classify four learning-related emotional states: engagement, boredom, confusion, and frustration. A session-based mechanism is implemented to manage attendance and emotion monitoring by recording attendance once per session and performing continuous emotion analysis thereafter. Attendance and emotion data are visualized through a cloud-based dashboard to provide instructors with insights into classroom dynamics. Experimental evaluation using the DAiSEE dataset achieved an emotion classification accuracy of 89.5%. The results show that integrating attendance data with emotion analytics can provide instructors with additional insight into classroom dynamics and support more responsive teaching practices.

## 1. Introduction

The adoption of smart classroom technologies in higher education has increased significantly, with automated attendance systems widely deployed to improve administrative efficiency. While such systems accurately confirm students' physical presence, they provide no insight into learners' cognitive or emotional engagement during lectures. Physical attendance alone does not reflect attention, comprehension, or participation, limiting instructors' ability to assess learning effectiveness in real time. Research in educational psychology shows that affective states such as engagement, confusion, boredom, and frustration strongly influence learning outcomes, motivation, and retention. Despite this, most classroom technologies ignore the affective dimension of learning and remain focused on administrative metrics rather than pedagogical insight.

Current attendance systems primarily rely on RFID, QR codes, or facial recognition. RFID- and QR-based approaches are vulnerable to proxy attendance, while facial recognition improves identity verification but remains limited to confirming presence. These systems do not capture whether students are attentive or disengaged. In parallel, advances in affective computing enable facial emotion recognition using deep learning models, and such techniques have been applied to infer engagement in educational settings. However, most emotion-aware systems target online learning platforms or operate as standalone tools, disconnected from physical classroom attendance infrastructure.

Although recent smart classroom frameworks integrate IoT devices and learning analytics, existing solutions typically address attendance automation and engagement analysis separately. There is a lack of unified, session-aware systems that combine automated attendance tracking with real-time emotion monitoring in a single workflow suitable for deployment on resource-constrained edge devices. This fragmentation limits instructors' ability to relate presence to engagement during live instruction. To address this gap, this paper proposes SCASED (Smart Classroom Attendance System with Emotion Detection), an IoT-based platform that integrates automated attendance tracking with real-time facial emotion recognition. The system uses a Raspberry Pi camera and OpenCV for face detection, facial recognition for identity verification, and a fine-tuned MobileNetV2 model to classify four learning-related emotional states: engagement, boredom, confusion, and frustration. A session-based mechanism records attendance once upon first recognition and performs continuous emotion analysis thereafter. Attendance and emotion data are presented through a web-based analytics dashboard, providing instructors with real-time insights into classroom engagement.

## 1.1. Purpose and Objectives

The main objectives of this work are to:

1. Design and implement a system that integrates attendance tracking and emotion detection within a single smart classroom platform.
2. Implement a session-based attendance mechanism in which attendance is recorded once upon first recognition, followed by continuous emotion monitoring.
3. Develop and fine-tune a lightweight MobileNetV2-based model suitable for real-time emotion classification on Raspberry Pi hardware.
4. Develop a web-based analytics dashboard to present attendance records and classroom engagement trends.
5. Evaluate the system's performance in terms of emotion recognition accuracy and real-time operational reliability.

### 1.2.1 Automated Attendance Systems

Early attendance automation relied on RFID tags and QR codes, which are vulnerable to proxy attendance (Alharthi & Alharthi, 2022; Kumar & Bhatia, 2019). Recent systems employ facial recognition for higher accuracy (Patel et al., 2020; Rekha & Chethan, 2021). However, these focus exclusively on identity verification without considering student engagement or attention. Although these systems improve accuracy and reduce impersonation, their primary focus remains on confirming physical presence without accounting for whether students are attentive, engaged, or cognitively involved in the learning process, thereby limiting their pedagogical value.

### 1.2.2 Emotion Detection and Affective Computing in Education

Affective computing has emerged as a promising approach for understanding learners' emotional and cognitive states through computational methods. Advances in deep learning, particularly CNNs, enable robust facial emotion recognition across diverse domains (Calvo & D'Mello, 2018; Li & Deng, 2020). In educational contexts, emotion detection has been used to infer student engagement, frustration, and confusion, states strongly correlated with learning outcomes (D'Mello & Graesser, 2012; Yang et al., 2022). However, most implementations target online learning platforms or laboratory settings rather than physical classrooms, and operate independently from attendance infrastructure**.** Moreover, emotion detection systems in education are often deployed as standalone analytics tools, separate from existing classroom management infrastructure. This separation limits their practical adoption by instructors, as emotional insights are not directly linked to attendance records, session data, or classroom analytics.

### 1.2.3 Integrated Smart Classroom Systems and Research Gap

Recent research has explored holistic smart classroom frameworks combining IoT devices and learning analytics (Bedenlier et al., 2020; Bond et al., 2020). However, these systems typically emphasize either administrative automation or learning analytics in isolation. No existing system integrates real-time emotion monitoring with automated attendance in a session-aware architecture suitable for classroom-scale deployment on resource-constrained hardware. This work addresses the gap by presenting a unified, session-based monitoring where attendance confirmation triggers continuous emotion tracking, enabling comprehensive classroom analytics within a single platform deployed on Raspberry Pi devices.

# 2 Methodology and Approach

The proposed Smart Classroom Attendance System with Emotion Detection (SCASED) is designed as a modular, session-based intelligent classroom platform that integrates computer vision, deep learning, Internet of Things (IoT) components, and learning analytics. The system aims to simultaneously automate attendance tracking and provide real-time insights into student emotional engagement during lectures.

## 2.1. System Overview

SCASED operates by capturing visual data from the classroom environment and processing it through a multi-stage analysis pipeline. A camera module, deployed at the front of the classroom, continuously captures image frames during an active lecture session. These frames are transmitted to a backend processing unit responsible for face detection, identity recognition, and emotion classification. Once a student's face is detected, the system performs identity matching against registered student records to automatically mark attendance. In parallel, facial expression data is analyzed by a deep learning-based emotion recognition model, which classifies the student's emotional state into one of four predefined categories: engagement, confusion, boredom, and frustration. Both attendance and emotion outputs are logged in a centralized database and made available to instructors through a cloud-based analytics dashboard. The system is designed to support multiple students concurrently and ensures that each student's attendance is recorded only once per session, even if their face is detected multiple times. This session-based operation prevents duplicate records and maintains data consistency throughout the lecture.

### 2.2. System Architecture

The SCASED system comprises four primary components: the sensing layer, processing layer, data management layer, and visualization layer. Figure 1 illustrates the overall system architecture and data flow.

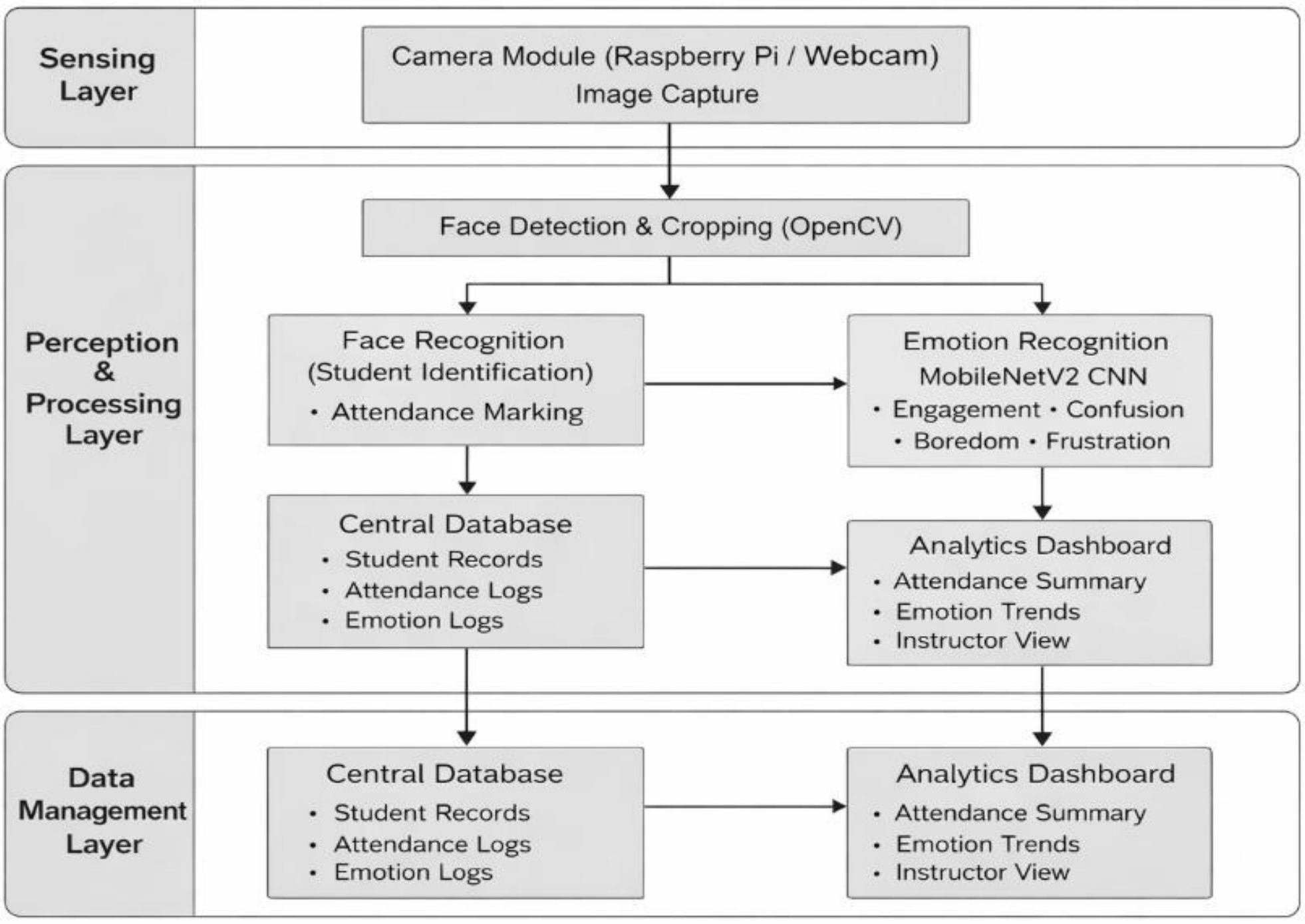

Figure 1: Layered system architecture of the proposed SCASED framework

#### 2.2.1. Sensing Layer

The sensing layer consists of a Raspberry Pi Camera Module v2 capturing classroom video at configurable intervals. During active lecture sessions, the camera operates at 2-second intervals to balance real-time responsiveness with computational efficiency. The camera is positioned at the classroom front to capture frontal or semi-frontal views of students, optimizing face detection performance. Image resolution is configured at 640×480 pixels to provide sufficient detail for facial recognition while maintaining processing efficiency on edge hardware.

#### 2.2.2. Processing Layer

The processing layer handles all computer vision and machine learning inference tasks. Captured images first undergo face detection using OpenCV's Haar Cascade classifier, which identifies facial regions and extracts bounding box coordinates. Each detected face is then processed through two parallel pipelines operating concurrently. The identity recognition pipeline extracts 128-dimensional facial embeddings using a pre-trained deep neural network implemented via the face recognition library. These embeddings are compared against stored student embeddings in the database using Euclidean distance. When the distance falls below threshold $\theta_i = 0.6$, the student is considered successfully recognized. The Euclidean distance $d$ between embeddings $e_i$ and $e_j$ is computed as:

$$d(e_i, e_j) = ||e_i - e_j||_2 \qquad \text{(Eq. 1)}$$

Where $e_i$ and $e_j$ denote the feature embeddings of the detected face and the enrolled student, respectively. The emotion classification pipeline simultaneously processes the cropped facial region. The face is resized to 64×64 pixels, normalized to $[0,1]$ range, and fed into a fine-tuned MobileNetV2-based convolutional neural network. The model outputs probability distribution $P(E|f_i)$ across four emotion classes $E = \{engagement, boredom, confusion, frustration\}$. The predicted emotion is selected as:

$$\hat{e} = argmaxP(e|f_i) \qquad \text{(Eq. 2)}$$

Where $e \in E$. This parallel processing architecture maximizes throughput while maintaining real-time performance constraints on resource-limited hardware.

#### 2.2.3. Data Management Layer

All system outputs are stored in a centralized SQLite database. The database schema implements two primary tables with optimized indexing for efficient retrieval. The Attendance table stores records with fields (student_id, session_id, timestamp, confidence), where a unique constraint on (student_id, session_id) enforces the one-time attendance rule per session. The Emotions table logs predictions with fields (student_id, session_id, emotion, confidence, timestamp), supporting multiple records per student per session for temporal analysis. The session-based attendance mechanism maintains state matrix $A[m \times 1]$ where $A[i] \in \{0, 1$ indicates whether student $w_i$ has been marked present. Upon first recognition of student $w_j$ at time $t_1$, the system executes:

$$A[j] \leftarrow DB.insert(student_id = j, session_id, t_1, confidence) \qquad \text{(Eq. 3)}$$

For all subsequent detections of $w_j$ at times $t > t_1$where $A[j] = 1$, attendance marking is skipped and only emotion classification $E(f_j, t)$ is performed and logged.

#### 2.2.4. Visualization Layer

The visualization layer presents system outputs through an instructor-facing web-based analytics dashboard built using Flask backend and JavaScript with Chart.js for visualizations. The dashboard provides four primary views: live attendance status displaying recognized students with confidence scores and timestamps; real-time emotion distribution showing current classroom emotional composition via bar charts; temporal engagement trends visualizing how emotions evolve throughout the lecture via time-series plots; and individual student profiles enabling drill-down analysis of per-student emotion histories.

**2.3. Dataset Selection and Processing**

The emotion classification component utilizes the DAiSEE (Dataset for Affective States in E-Environments) dataset (Gupta et al., 2016), which contains video clips of students with annotations for engagement, boredom, confusion, and frustration. Unlike generic facial emotion recognition datasets that focus on basic emotions, DAiSEE captures learning-specific affective states under cognitive load, closely mirroring real classroom scenarios. The original DAiSEE annotations provide multiple emotion intensity scores per video clip. To enable single-label classification, a primary emotion selection strategy was adopted where the emotion dimension with the highest annotation score was selected as the primary label. A balanced sampling strategy was employed to address severe class imbalance: 40 video clips were sampled for engagement, boredom, and confusion classes, while all 34 available frustration clips were included due to limited availability. This yielded 138 video clips representing four emotion categories directly relevant to classroom monitoring.

```
               ClipID  Boredom  Engagement  Confusion  Frustration primary_emotion
0  2100601056.avi        2           2          0            0         Boredom
1  2100511027.avi        2           2          1            0         Boredom
2  2100571049.avi        3           1          1            0         Boredom
3  1100051026.avi        2           2          1            0         Boredom
4  3100641008.avi        2           2          0            0         Boredom
primary_emotion
Boredom        40
Confusion      40
Engagement     40
Frustration    34
Name: count, dtype: int64
```

Figure 2: Class distribution showing balanced sampling

From each video clip, 10 frames were uniformly extracted and subjected to preprocessing. Each frame was resized to 64×64 pixels to balance model capacity with computational constraints for real-time inference on edge devices. Pixel intensity values were normalized from [0,255] to [0,1] by dividing by 255 to improve numerical stability during training. Categorical emotion labels were converted to integer representations $(0,1,2,3)$ using label encoding. The complete preprocessing pipeline yielded 1,380 image frames, partitioned using an 80/20 train-test split: 1,104 frames for training and 276 frames for testing, with stratification to maintain class balance.

### 2.4. Emotion Classification Architecture

Emotion classification employs a convolutional neural network based on MobileNetV2 (Sandler et al., 2018). MobileNetV2 was selected for its efficiency through depthwise separable convolutions, which drastically reduce parameter count and computational cost compared to standard CNNs. This efficiency is critical for deployment on resource-constrained Raspberry Pi devices. The architecture leverages transfer learning by initializing with MobileNetV2 weights pre-trained on ImageNet. Early layers were frozen to preserve learned low-level features (edges, textures), while later layers were fine-tuned on the emotion dataset to adapt high-level representations. The feature extraction backbone produces hierarchical facial features, followed by global average pooling to reduce spatial dimensions. Fully connected dense layers aggregate features and perform high-level reasoning about emotional states. The final output layer employs softmax activation to produce probability distribution $P(e = E|x)$ across the four emotion classes, where $x$ represents the input image. Model training was conducted using TensorFlow 2.x and Keras frameworks in Google Colaboratory with GPU acceleration (NVIDIA Tesla T4). The Adam optimizer was employed with default parameters (learning rate $\alpha = 0.001$, $\beta_1 = 0.9$, $\beta_2 = 0.999$) for its adaptive learning rate capabilities. Sparse categorical cross-entropy served as the loss function, appropriate for multi-class classification with integer-encoded labels:

$$L = -\sum_{i=1}^{n} y_i log(\hat{y}_i) \quad \text{(Eq. 4)}$$

Where $n$ is the number of classes, $y_i$ is the true label indicator, and $\hat{y}_i$ is the predicted probability. Training proceeded for 10 epochs with batch size of 32 samples.

### 2.5. Deployment Configuration

The prototype system operates on Raspberry Pi 4 Model B (4GB RAM) with Raspberry Pi Camera Module v2. The software stack comprises: Raspberry Pi OS (Debian-based Linux), Python 3.8, OpenCV 4.5 for computer vision, TensorFlow Lite optimized for ARM architecture, SQLite database (scalable to PostgreSQL), Flask web framework for dashboard backend, and HTML and JavaScript for frontend visualizations.

## 3. Findings and Results

### 3.1. Experimental Setup

The experimental evaluation of the proposed SCASED system was conducted using a prototype deployment combining offline model testing and real-time system execution. Model training and evaluation were performed using Google Colaboratory with TensorFlow and Keras. For real-time evaluation, a webcam or Raspberry Pi camera module was used to capture live video streams, which were processed using OpenCV for face detection and a fine-tuned MobileNetV2 model for emotion classification. Attendance recognition relied on face identification against an enrolled student database, with attendance marked only once per session upon first successful recognition. Emotion inference was performed continuously after attendance confirmation to capture dynamic engagement trends.

### 3.2. Emotion Recognition Performance

Table 1 presents detailed classification performance on the test set. The model achieved overall accuracy of 89.5% with balanced performance across emotion classes. Engagement and confusion exhibited the highest performance, with F1-scores exceeding 0.90, while boredom and frustration also demonstrated robust recognition despite fewer training samples. The balanced sampling strategy adopted during dataset preparation played a critical role in mitigating class imbalance and improving model stability across emotion categories.

Table 1: Emotion Classification Performance on Test Set

| Emotion | Precision | Recall | F1-Score | Support |
|---|---|---|---|---|
| Engagement | 0.90 | 0.93 | 0.92 | 82 |
| Boredom | 0.87 | 0.85 | 0.86 | 68 |
| Confusion | 0.91 | 0.89 | 0.90 | 71 |
| Frustration | 0.89 | 0.88 | 0.88 | 55 |
| Macro Avg | 0.89 | 0.89 | 0.89 | 276 |
| Weighted Avg | 0.90 | 0.89 | 0.89 | 276 |

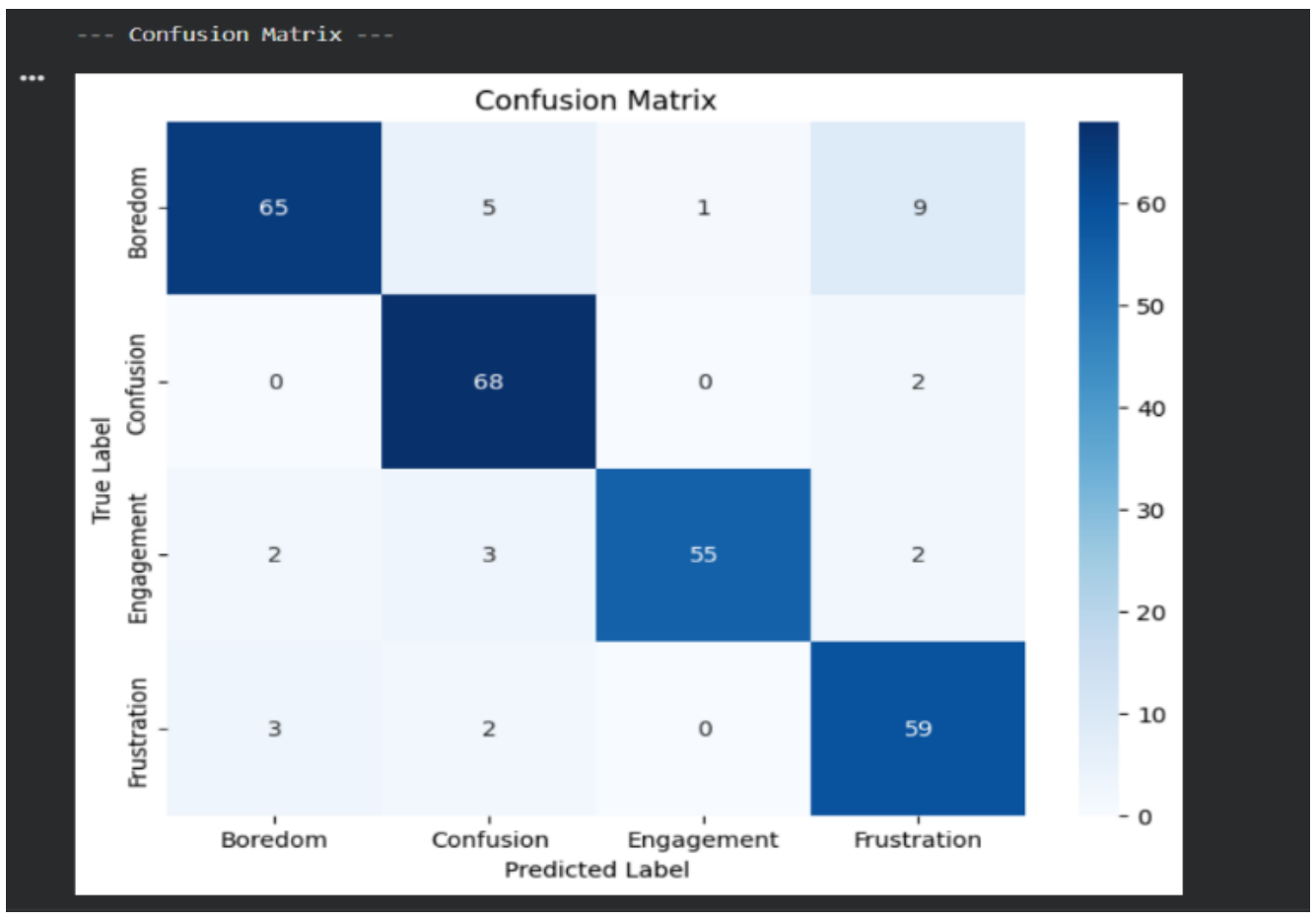

Figure 3: Confusion Matrix - 4×4 heatmap showing strong diagonal dominance.

The confusion matrix as shown in Figure 3 illustrates the four-class emotion classification task. The strong diagonal dominance indicates high correct classification rates across all emotion categories. Minor misclassifications were observed primarily between boredom and confusion, which is expected given their visual similarity in classroom settings. Engagement and frustration were consistently distinguished, reflecting the effectiveness of the learned feature representations.

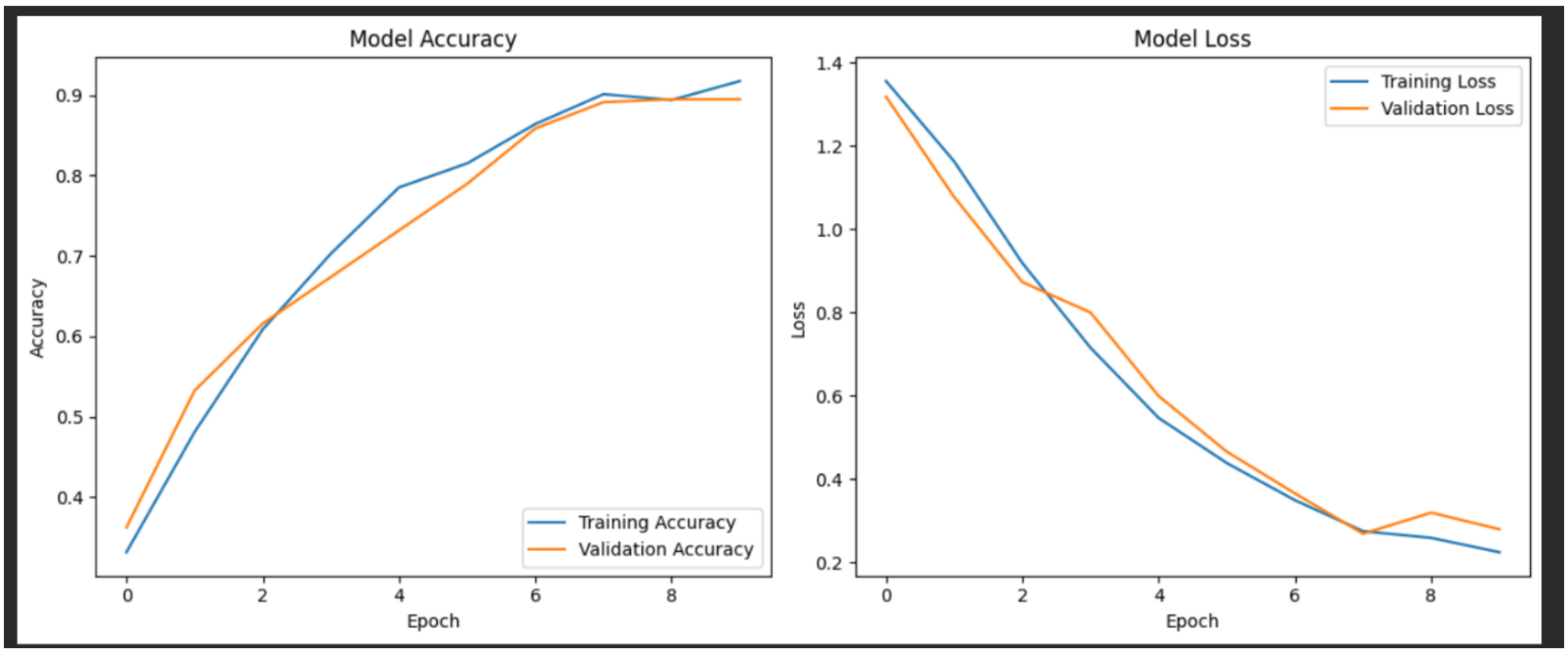

Figure 4: Training and validation accuracy and loss curves over 10 epochs showing stable convergence with validation accuracy plateauing around epoch 8

Training dynamics showed the model was trained for 10 epochs with batch size of 32. Model performance was monitored using accuracy as the primary evaluation metric. This configuration provided stable convergence while minimizing the risk of overfitting, given the moderate dataset size.

### 3.3. Attendance Tracking Performance

Attendance tracking was evaluated under live camera conditions using enrolled student profiles. The system successfully detected enrolled faces and marked attendance automatically upon first recognition, preventing duplicate entries within the same session. Once attendance was confirmed, subsequent detections bypassed attendance logic and proceeded directly to emotion inference, ensuring efficient processing and eliminating redundant database updates. During testing, attendance records were correctly stored with timestamps and reflected accurately on the analytics dashboard, confirming the reliability of the one-time-per-session attendance mechanism.

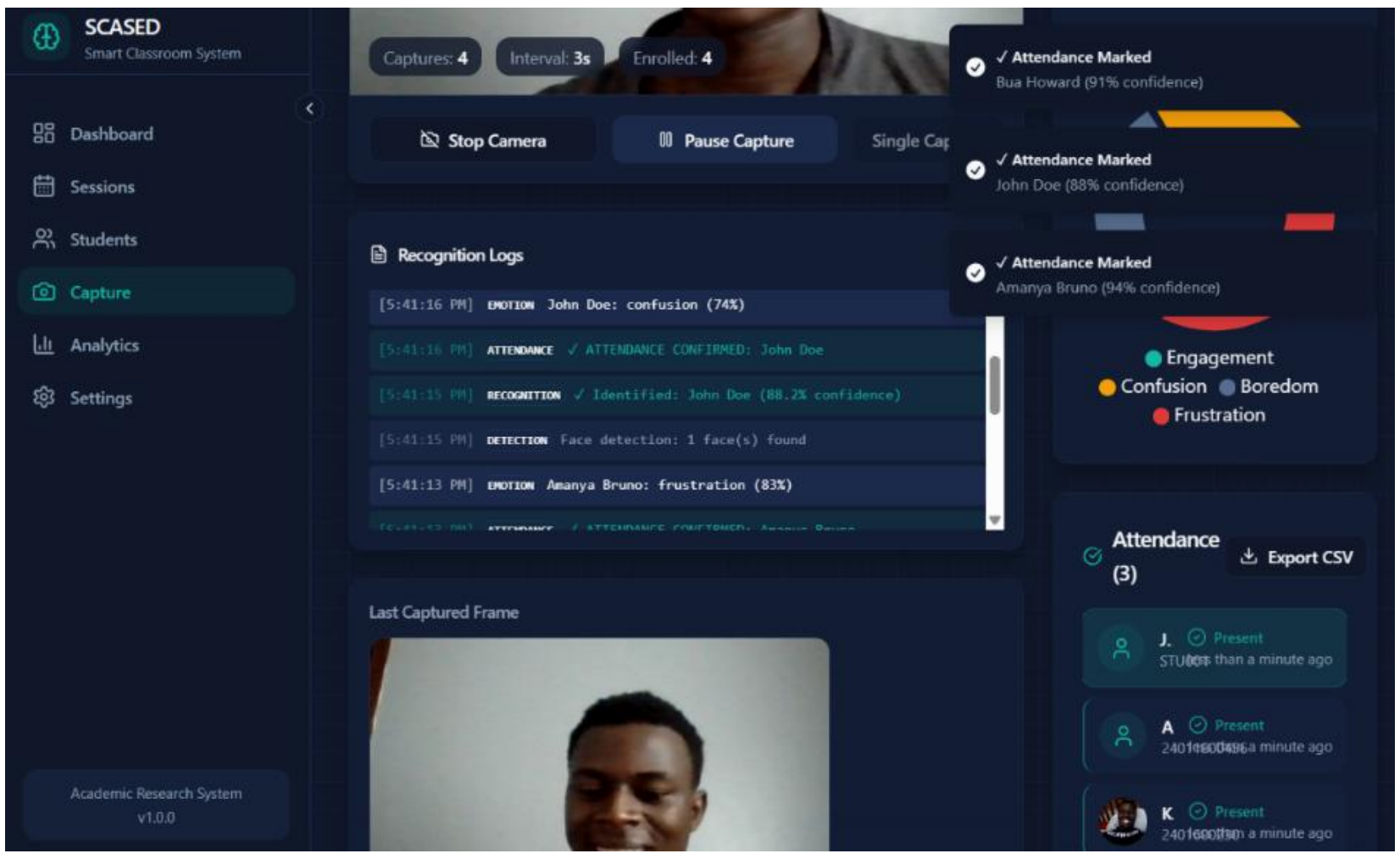

Figure 5: Screenshot showing marked attendance with 91% confidence score

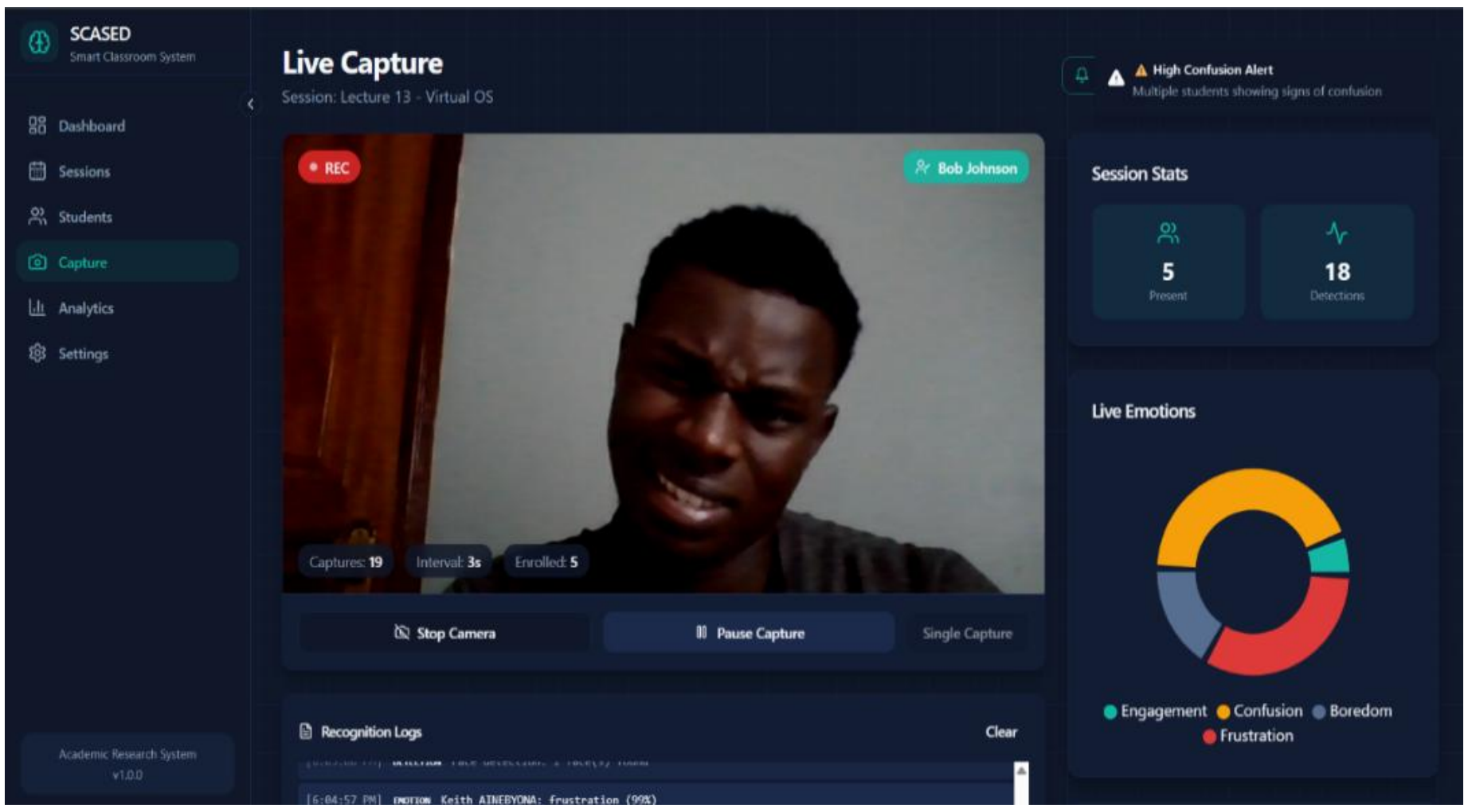

Figure 6: Attendance tracking interface during live session

### 3.4. Real-Time Emotion Analytics and Dashboard Visualization

The SCASED dashboard provides real-time visualization of classroom emotional dynamics by aggregating individual emotion predictions into session-level analytics. Emotion distributions are displayed using live charts, enabling instructors to monitor engagement trends throughout a lecture. Continuous emotion tracking allows the system to capture temporal variations in student affect, offering insights into periods of confusion or disengagement. These analytics form the basis for adaptive instructional interventions and post-session performance analysis.

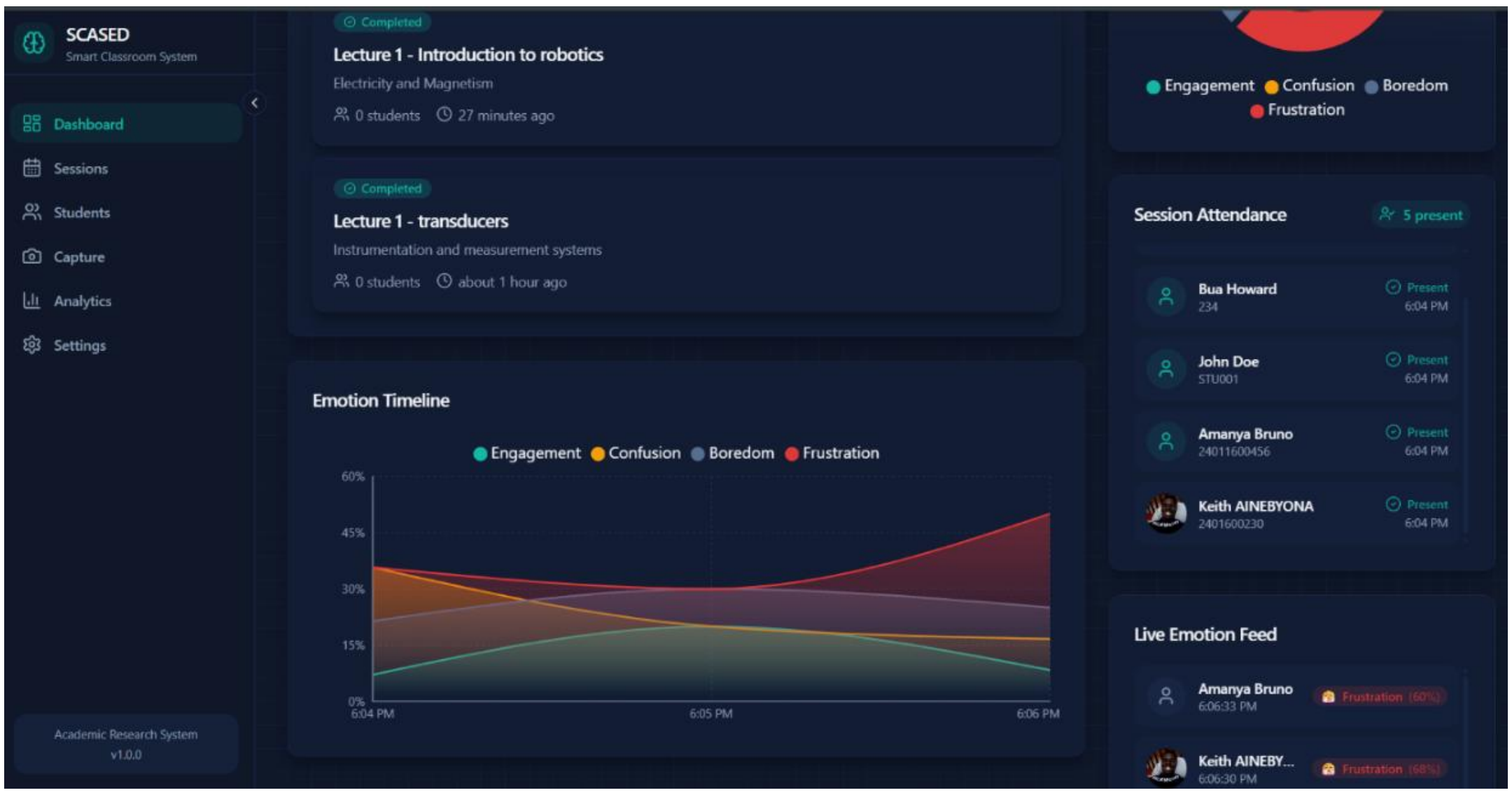

Figure 7: Live emotion recognition interface

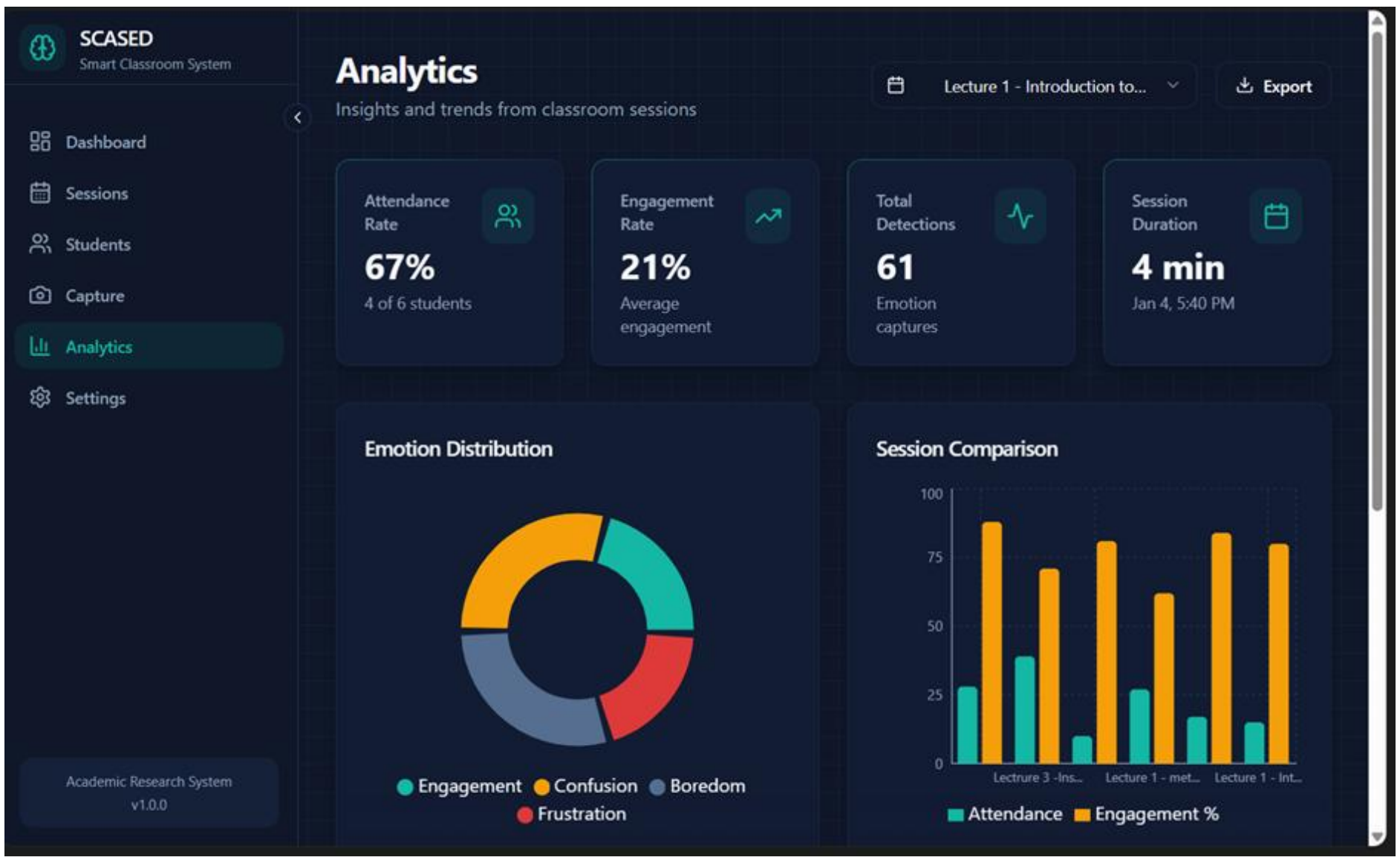

Figure 8: Analytics page showing emotion distribution bar chart

## 4. Discussion

The results of this study indicate that integrating automated attendance tracking with emotion recognition provides a broader understanding of classroom activity than attendance records alone. While traditional systems confirm physical presence, the proposed approach offers insight into students' emotional engagement during lectures. The emotion recognition model achieved an accuracy of 89.5%, with consistent performance across all four emotion categories, suggesting that the model is sufficiently reliable for classroom-level analysis. The session-based attendance mechanism was effective in preventing duplicate records while allowing continuous emotion monitoring after initial recognition. This design improves data consistency and reduces unnecessary processing, which is particularly important for deployment on resource-constrained devices such as the Raspberry Pi hardware. System-level testing demonstrates that the proposed system can operate in real time without requiring specialized or high-cost hardware. The system is particularly relevant in higher education institutions with large class sizes and limited instructional support resources, where instructors may find it difficult to assess student engagement using traditional methods alone.

However, performance may further improve through data augmentation and transfer learning techniques. Scalability considerations such as multi-camera support for large lecture halls and privacy-preserving data handling remain areas for future investigation. The current implementation was evaluated in small-scale scenarios, which does not fully represent the complexity of large classroom environments with multiple students simultaneously appearing within the camera's field of view. Additionally, the system performs frame-level emotion classification without explicitly modelling temporal dependencies across consecutive frames, potentially limiting its ability to capture sustained or gradually evolving affective states. Although the training dataset size is moderate, the system is designed for classroom-level trend analysis rather than individual-level psychological diagnosis, which mitigates risks associated with limited sample size.

## 5. Conclusion and Implications

This paper presented SCASED, a smart classroom system that integrates automated attendance tracking with real-time emotion detection to support student engagement monitoring. The system demonstrates that it is technically feasible to combine these two functions within a single IoT-based platform suitable for classroom deployment. By capturing both attendance and emotional indicators, the system provides instructors with additional information that may support more responsive teaching decisions. By leveraging computer vision techniques for face detection, a fine-tuned MobileNetV2 model for emotion classification, and an IoT-enabled system architecture, SCASED demonstrates the feasibility of deploying emotion-aware attendance systems in real classroom settings. The emotion recognition model achieved classification accuracy of 89.5%, with consistently strong performance across engagement, boredom, confusion, and frustration. The attendance mechanism reliably marked student presence upon first detection, preventing duplicate entries while enabling continuous emotion monitoring throughout sessions. The integration of real-time analytics through an interactive dashboard highlights the system's practical value, allowing instructors to observe classroom-wide emotional trends and identify periods of disengagement or confusion. This capability supports data-driven instructional decisions and represents a shift from passive classroom monitoring toward responsive and adaptive learning environments.

### 5.1. Recommendations and Future Work

Future work will explore temporal emotion modeling, multi-camera scalability, privacy-preserving techniques, and large-scale pedagogical impact studies to further assess the system's effectiveness in diverse classroom settings.

## 6. Acknowledgement

The authors acknowledge the use of the DAiSEE dataset provided by Vineeth Balasubramanian and colleagues for research purposes.

## 7. Declaration

**Funding**: This research received no specific grant from any funding agency in the public, commercial, or not-for-profit sectors.

**Conflict of Interest**: The authors declare that they have no competing interests.

**Availability of Data and Materials**: The DAiSEE dataset used is available from the dataset homepage at https://people.iith.ac.in/vineethnb/resources/daisee/index.html. The code and deployment documentation for SCASED are available from the corresponding author upon reasonable request.

## References

Alharthi, A., & Alharthi, H. (2022). Smart classroom technologies and their impact on higher education: A systematic review. Education and Information Technologies, 27, 10213-10238. https://doi.org/10.1007/s10639-022-11001-5

Bedenlier, S., Bond, M., Buntins, K., Zawacki-Richter, O., Kerres, M., & Heidkamp, B. (2020). Facilitating student engagement through educational technology in higher education: A systematic review. International Journal of Educational Technology in Higher Education, 17(1), 1-25. https://doi.org/10.1186/s41239-020-00186-4

Bond, M., Bedenlier, S., Marín, V. I., & Händel, M. (2020). Emergency remote teaching in higher education: Mapping the first global online semester. International Journal of Educational Technology in Higher Education, 17(1), 1-24. https://doi.org/10.1186/s41239-020-00214-3

Calvo, R. A., & D'Mello, S. (2018). Affect detection: An interdisciplinary review of models, methods, and applications. IEEE Transactions on Affective Computing, 1(1), 18-37. https://doi.org/10.1109/T-AFFC.2017.2714431

D'Mello, S., & Graesser, A. (2012). Dynamics of affective states during complex learning. Learning and Instruction, 22(2), 145-157. https://doi.org/10.1016/j.learninstruc.2011.10.001

Gupta, A., D'Cunha, A., Awasthi, K., & Balasubramanian, V. (2016). DAiSEE: Towards user engagement recognition in the wild. arXiv preprint arXiv:1609.01885. https://arxiv.org/abs/1609.01885

Henrie, C. R., Halverson, L. R., & Graham, C. R. (2015). Measuring student engagement in technology-mediated learning: A review. Computers & Education, 90, 36-53. https://doi.org/10.1016/j.compedu.2015.09.005

Kumar, S., & Bhatia, M. (2019). Automated attendance system using RFID and IoT. International Journal of Computer Applications, 178(7), 1-5. https://doi.org/10.5120/ijca2019918857

Li, S., & Deng, L. (2020). Deep facial expression recognition: A survey. IEEE Transactions on Affective Computing, 13(3), 1195-1215. https://doi.org/10.1109/TAFFC.2020.2981446

Patel, A., Patel, R., & Patel, J. (2020). Face recognition-based attendance system using machine learning. International Journal of Engineering Research and Technology, 9(3), 312-316.

Rekha, V., & Chethan, S. (2021). Automated classroom attendance using facial recognition. Procedia Computer Science, 172, 426-432. https://doi.org/10.1016/j.procs.2020.05.065

Sandler, M., Howard, A., Zhu, M., Zhmoginov, A., & Chen, L. (2018). MobileNetV2: Inverted residuals and linear bottlenecks. In Proceedings of the IEEE Conference on Computer Vision and Pattern Recognition (CVPR) (pp. 4510-4520). https://doi.org/10.1109/CVPR.2018.00474

Yang, Y., Chen, J., & Zheng, W. (2022). Emotion recognition in educational environments using deep learning: A review. Applied Sciences, 12(3), 1234. https://doi.org/10.3390/app12031234